%% file: main.tex
\documentclass{article}
\usepackage[utf8]{inputenc}
\usepackage{amsmath}
\usepackage{amssymb}
\usepackage{graphicx}
\usepackage{caption}
\usepackage{subcaption}
\usepackage{url}
\usepackage{booktabs}
\usepackage{textcomp}
\usepackage{float}

\begin{document}

\title{Calibrated Uncertainty for Molecular Property Prediction using Ensembles of Message Passing Neural Networks}

\author{
    Jonas Busk$^{1,*}$ \and 
    Peter Bj{\o}rn J{\o}rgensen$^1$ \and
    Arghya Bhowmik$^1$ \and 
    Mikkel N. Schmidt$^2$ \and
    Ole Winther$^{2,3,4}$ \and
    Tejs Vegge$^1$
}

\date{\small{
    $^{1}$Department for Energy Conversion and Storage, Technical University of Denmark, Lyngby, Denmark \\
    $^{2}$Department of Applied Mathematics and Computer Science, Technical University of Denmark, Lyngby, Denmark \\
    $^{3}$Center for Genomic Medicine, Rigshospitalet, Copenhagen University Hospital, Copenhagen, Denmark \\
    $^{4}$Bioinformatics Centre, Department of Biology, University of Copenhagen, Copenhagen, Denmark \\
    $^{*}$Corresponding author: Jonas Busk, jbusk@dtu.dk
}}

\maketitle

\begin{abstract}
Data-driven methods based on machine learning have the potential to accelerate computational analysis of atomic structures.
In this context, reliable uncertainty estimates are important for assessing confidence in predictions and enabling decision making.
However, machine learning models can produce badly calibrated uncertainty estimates and it is therefore crucial to detect and handle uncertainty carefully.
In this work we extend a message passing neural network designed specifically for predicting properties of molecules and materials with a calibrated probabilistic predictive distribution.
The method presented in this paper differs from previous work by considering both aleatoric and epistemic uncertainty in a unified framework, and by recalibrating the predictive distribution on unseen data.
Through computer experiments, we show that our approach results in accurate models for predicting molecular formation energies with well calibrated uncertainty in and out of the training data distribution on two public molecular benchmark datasets, QM9 and PC9.
The proposed method provides a general framework for training and evaluating neural network ensemble models that are able to produce accurate predictions of properties of molecules with well calibrated uncertainty estimates.
\end{abstract}

\hspace{.5em}

\noindent{\it Keywords\/}: Molecular property prediction, machine learning potential, uncertainty quantification, uncertainty calibration, message passing neural network, graph neural network, ensemble model.

\section{Introduction}

Autonomous high-throughput computational analysis of atomic structures has the potential to speed up the discovery of novel materials and chemical reactions dramatically with applications in a wide range of research areas including biotechnology and conversion and storage of renewable energy.
This process can be enabled and accelerated by data-driven methods based on machine learning that are generally less computationally demanding than traditional quantum mechanical methods such as density functional theory (DFT)~\cite{dral2020, lilienfeld2020}.
In this context, reliable uncertainty estimates are important to assess confidence in predictions and thereby enable decision making and automation~\cite{peterson2017, tran2020}.
In recent years, graph-based models such as message passing neural networks (MPNNs), that operate on atomic structures represented as graphs, have shown impressive capabilities at predicting properties of molecules and materials with high accuracy~\cite{messagepassing}. 
However, deep neural networks are known to produce badly calibrated uncertainty estimates on regression tasks~\cite{kuleshov2018, song2019, levi2020, kahle2021},
especially outside the training data distribution, which can lead to sub-optimal or incorrect results.
Because chemical space is too vast to represent in any training dataset~\cite{gomez-bombarelli, polishchuk2013}, it is crucial to quantify and handle predictive uncertainty carefully in this setting, for example by falling back to more accurate but computationally demanding methods like DFT when uncertainty is high~\cite{fornari2020}.
Consequently, predictive algorithms that express reliable probabilistic uncertainty estimates can help identify problematic instances and enable the design of new robust workflows and applications in computational materials science, such as active learning and autonomous high throughput screening~\cite{bolle2020, eyke2020, hie2020, soleimany2021}.

When quantifying uncertainty it is often useful to distinguish between \emph{epistemic} and \emph{aleatoric} uncertainty~\cite{uncertainty_in_ml, gal2017}.
Epistemic uncertainty, also known as systematic uncertainty, arises from the model's inability to fit the data distribution and can in principle be reduced by observing more data or improving the model.
Aleatoric uncertainty, also known as statistical uncertainty, on the other hand comes from inherent noise in the data and can therefore not be reduced by observing more data.
When the aleatoric uncertainty is constant across all observations it is called \emph{homoscedastic} aleatoric uncertainty and is often not modelled explicitly.
If the aleatoric uncertainty depends on the input, and thus varies across the data distribution, it is called \emph{heteroscedastic} aleatoric uncertainty and can be estimated from the data by explicitly including it in the model.
Thus epistemic uncertainty is important for understanding when predictions are reliable and aleatoric uncertainty captures noise in the data.
Consequently, it is necessary to consider both types of uncertainty to obtain a complete picture of the predictive uncertainty and to achieve well calibrated uncertainty estimates in and out of the training data distribution.


Uncertainty quantification for property prediction of atomic structures with graph neural networks has received increasing interest in recent research.
Scalia et al.~\cite{scalia2020} evaluated and compared scalable uncertainty estimation methods based on graph neural networks for molecular property prediction and found that deep ensembles~\cite{deepensemble} and bootstrapping consistently outperformed Monte Carlo Dropout~\cite{gal2016} on multiple public benchmark datasets in terms of error and uncertainty calibration.
Hirschfeld et al.~\cite{hirschfeld2020} compared several uncertainty quantification methods, including graph neural networks, on four molecular benchmark datasets, but did not find a method that performed consistently well across datasets.
Tran et al.~\cite{tran2020} highlighted the importance of predictive uncertainty in materials screening applications and reviewed methods for uncertainty quantification and procedures for evaluating the quality of uncertainty estimates including accuracy, calibration and sharpness.
Soleimany et al.~\cite{soleimany2021} evaluated deep evidential regression as a method of uncertainty quantification for molecular property prediction and demonstrated their approach in active learning and virtual screening applications.
Nigam et al.~\cite{nigam2021} provided an extensive overview of different sources of uncertainty in molecular property prediction in the context of drug discovery, many of which are also relevant in materials science, and described the importance and perspectives of having good uncertainty estimates in data driven decision making. 
Related work has studied the use of Gaussian process regression models for molecular property prediction~\cite{musil2019} and molecular dynamics~\cite{imbalzano2021}.
The method presented in this paper differs from the previous work by considering both aleatoric and epistemic uncertainty, and by recalibrating the predictive distribution to obtain more accurate uncertainty estimates on unseen data.

The main contribution of this paper is a complete framework for training and evaluating neural network ensemble models that are able to produce accurate predictions of properties of molecules with well calibrated uncertainty estimates in and out of the training data distribution.
Specifically, we extend a message passing neural network regression model designed for predicting properties of molecules and materials~\cite{jorgensen} with a probabilistic predictive distribution and consider a deep ensemble of models~\cite{deepensemble} to express aleatoric and epistemic uncertainty about predictions of molecular formation energies.
The uncalibrated predictive distribution is recalibrated \emph{post hoc} to fit the error distribution on unseen data to address model overconfidence from training and the expected reduction in error from using an ensemble approximation.
Through computer experiments we show that our approach results in accurate and well calibrated models on two public benchmark datasets for molecular property prediction, QM9~\cite{qm9} and PC9~\cite{pc9}, and additionally that out of distribution predictions are also well calibrated when training on QM9 and testing on the more diverse PC9 dataset.

The rest of the paper is organised as follows.
The proposed method is described in detail in section~\ref{sec:method} and experiments and results are presented in section~\ref{sec:results}.
The main findings and perspectives are discussed in section~\ref{sec:discussion} and finally we conclude in section~\ref{sec:conclusion}.

\section{Method}
\label{sec:method}

\subsection{Message passing neural network model}

In general, a message passing neural network (MPNN), as described in~\cite{messagepassing}, operates on a graph structure $g$ with node features $x_v$ and edge features $e_{vw}$, where $v$ and $w$ denote vertices in the graph.
A forward pass through the neural network consists of two phases: i) a message passing phase with $T$ interaction steps where messages are passed along the edges of the graph to update the internal graph embedding, and ii) a readout phase where an output value $\hat y$ is computed from the final graph embedding. 

We base our work on the SchNet with edge updates MPNN model, which was previously introduced by the authors~\cite{jorgensen}. This model is in turn based on the popular SchNet model, that was designed specifically for predicting properties of molecules and materials~\cite{schnet}.
We refer the reader to the cited literature for specific details about this neural network architecture.
It is worth noting that the uncertainty quantification method proposed in the following sections does not depend on the particular choice of neural network model and can thus be adapted to use other models based on the specific application.

\subsection{Extended model with predictive uncertainty}

To capture both epistemic and heteroscedastic aleatoric uncertainty, we extend the MPNN described in the previous section by constructing a deep ensemble of neural networks~\cite{deepensemble} (without adversarial training) in the following way.
Given a regression task with a training dataset $\mathcal{D}=\{g_n,y_n\}_{n=1}^N$ consisting of $N$ datapoints with real-valued targets $y \in \mathbb{R}$,
we consider an ensemble of $M$ neural network models with parameters $\{\theta_m\}_{m=1}^M$,
each with probabilistic predictive distribution:
\begin{equation}
p_\theta(y|g) = \mathcal{N}\big( \mu_\theta(g), \sigma_\theta^2(g) \big) \, ,
\end{equation}
assuming a normal distribution of errors.
Each network is constructed with two outputs corresponding to the predicted mean $\mu_\theta(g)$ and variance $\sigma_\theta^2(g)$, where the latter represents the predicted heteroscedastic aleatoric uncertainty~\cite{nix}.
The predicted variance is constrained to be positive by passing the second network output through the softplus function, $\log(1+\exp(\cdot))$, and adding a small minimum variance for numerical stability (e.g. $10^{-6}$).

\subsection{Model training procedure}

Each network in the ensemble is initialized with random parameters and trained individually on the same training dataset using stochastic gradient descent to minimise the negative log likelihood (NLL) loss:
\begin{align}
\text{NLL}(\theta) &=
\frac{1}{N} \sum_{n=1}^N -\log p_\theta(y_n|g_n) \\
&= 
\frac{1}{N} \sum_{n=1}^N
\frac{1}{2} \bigg( \frac{1}{\sigma_\theta^2(g_n)}
\underbrace{\big( y_n-\mu_\theta(g_n)\big)^2}_{\text{squared error}}
+ \log \sigma_\theta^2(g_n) + \underbrace{\log2\pi}_{\text{constant}} \bigg)
\, .
\label{eq:normalnll}
\end{align}
The last term in equation~\ref{eq:normalnll} is constant since it does not depend on $\mu_\theta(g)$ or $\sigma_\theta^2(g)$ and can be ignored for the purpose of training the model.
Notice how for constant variance (homoscedastic uncertainty) this is equivalent to minimising the mean squared error (MSE) loss often used in regression.
Notice also how the predicted uncertainty acts as learned loss attenuation by letting examples with high predicted uncertainty have smaller impact on the total loss, while the $\log \sigma_\theta^2$ term discourages large uncertainties~\cite{gal2017}.

In practice, we found that training directly with NLL loss can be unstable because of interactions between the mean and variance output in the loss function.
To mitigate this, we initially train the mean output of the network before introducing the variance terms by interpolating from MSE to NLL loss:
\begin{equation}
\mathcal{L}(\theta) = \lambda \, \text{MSE}(\theta) + (1-\lambda) \, \text{NLL}(\theta) \, ,
\label{eq:loss_interpolation}
\end{equation}
where $\lambda$ is set to 1 for a number of warmup steps and then decreased linearly from 1 to 0 over a number of interpolation steps.
The resulting loss function is quite natural since the NLL loss includes the squared error term (see equation~\ref{eq:normalnll}) and as a result we found that model training becomes more stable and robust to outliers in the training data.
Additional measures exist to promote the stability of training variance networks~\cite{nix, amini, skafte}, but we found the method above to be sufficient in our experiments.

\subsection{Ensemble mixture}

To produce the ensemble predictive distribution $p_*(y|g)$ and capture epistemic uncertainty, we follow the approach of~\cite{deepensemble} and make an ensemble approximation by combining the predictions of the $M$ individual models as a uniformly-weighted mixture of normal distributions:
\begin{equation}
    p_*(y|g)=\frac{1}{M}\sum_{m=1}^M p_{\theta_m}(y|g) \, ,
\end{equation}
whose mean $\mu_*(g)$ and variance $\sigma_*^2(g)$ are given by the following expressions:
\begin{align}
\mu_*(g) &= \frac{1}{M} \sum_{m=1}^M \mu_{\theta_m}(g) \, ,
\\
\sigma_*^2(g) &= \frac{1}{M} \sum_{m=1}^M \big( \sigma_{\theta_m}^2(g) + \mu_{\theta_m}^2(g) \big) - \mu_*^2(g)
\\
&= \underbrace{\frac{1}{M} \sum_{m=1}^M \sigma_{\theta_m}^2(g)}_{\text{aleatoric uncertainty}} + \underbrace{\frac{1}{M} \sum_m \mu_{\theta_m}^2(g) - \mu_*^2(g)}_{\text{epistemic uncertainty}} \, .
\label{eq:ensemble_variance}
\end{align}
The variance of the ensemble predictive distribution represents the total predicted uncertainty and can be decomposed into aleatoric and epistemic uncertainty as shown in equation~\ref{eq:ensemble_variance} above.

\subsection{Uncertainty calibration and sharpness}

Intuitively, uncertainty calibration means there should be some kind of agreement between the predicted distribution and the empirical distribution~\cite{song2019}.
The concept of calibration has been studied extensively in the area of classification, where a classifier is said to be well calibrated
if the predicted class probability corresponds to the empirical probability that the instance belongs to that class~\cite{degroot1983, dawid1982, guo2017}.
In other words, the classifier is expected to correctly predict its error.
A few recent works have aimed to develop a corresponding definition of calibration in the area of regression~\cite{kuleshov2018, song2019, levi2020}.
Kuleshov et al.~\cite{kuleshov2018} propose that a model is well calibrated if the quantiles of the predicted distribution corresponds to the quantiles of the empirical distribution averaged over the data.
This approach is referred to as \emph{quantile-calibration} by Song et al.~\cite{song2019} who propose an alternative definition which they call \emph{distribution-calibration}, stating that a model is well calibrated if for all predictions with the same predictive distribution, the predictive distribution corresponds to the empirical distribution.
They proceed to show that if a model is distribution-calibrated it is also quantile-calibrated.
Levi et al.~\cite{levi2020} propose a definition where a model is well calibrated if the predicted uncertainty corresponds to the expected empirical error.
Following \cite{scalia2020}, we will refer to this as \emph{error-calibration} and we note that for any unbiased model with an expected error of zero, error-calibration corresponds exactly to distribution-calibration.
Based on these definitions, we find it useful and intuitive to interpret the predicted uncertainty as an indication of the expected error.

Assessing the quality of uncertainty estimates in regression tasks directly is not straight forward as the true uncertainties are generally unknown, but we can instead assess the uncertainty calibration by evaluating metrics derived from the definitions above~\cite{tran2020, kuleshov2018, song2019, levi2020, scalia2020}.
The NLL is the standard metric for evaluating the quality of probabilistic models by measuring the probability of observing the data given the predicted distribution.
However, in regression the NLL depends both on the predicted mean and variance (see equation~\ref{eq:normalnll}), and therefore it is useful to additionally evaluate the predicted uncertainty on its own.
To evaluate the error-calibration of a regression model we compare the predicted uncertainties to the corresponding empirical errors on unseen data.
In practice we sort examples by their predicted uncertainty, divide them into $K$ equal sized bins and compute the predicted root mean variance (RMV) and the empirical root mean squared error (RMSE) in each bin $k$.
Plotting RMV against RMSE shows if the predicted uncertainty corresponds to the empirical error in each bin on average and a straight diagonal line corresponding to the identity function indicates perfect error-calibration.
The error-calibration can be summarized by the expected normalized calibration error (ENCE), which is analogues to the expected calibration error (ECE) often used in classification~\cite{levi2020}:
\begin{equation}
    \text{ENCE} = \frac{1}{K}\sum_{k=1}^K \frac{|\text{RMV}_k-\text{RMSE}_k|}{\text{RMV}_k} \, .
\end{equation}
To additionally evaluate the quantile-calibration of a model, we compare the quantiles of the predictive distribution to the quantiles of the empirical distribution averaged over a set of unseen data~\cite{kuleshov2018}.
Plotting the predicted quantiles against the empirical quantiles shows if the predictive distribution corresponds to the empirical distribution on average and again a straight diagonal line corresponding to the identity function indicates perfect quantile-calibration.
The quantile-calibration can be summarised by the sum of squared errors (SSE) between the predicted and empirical quantiles.
To further evaluate the ability of a model to rank predictions by uncertainty with respect to error on unseen data, we sort predictions by uncertainty in decreasing order and plot the variation in error as we leave out the most uncertain predictions~\cite{soleimany2021, scalia2020}.
For a well calibrated model, we expect the error to decrease monotonically as the most uncertain predictions are omitted.
However, we do not expect a perfect ranking with respect to the errors since some highly uncertain predictions can still have small errors.

Calibration alone is not sufficient to ensure that individual uncertainty estimates are informative~\cite{tran2020, kuleshov2018, scalia2020, degroot1983}.
For example, a regression model that predicts constant uncertainty corresponding to its average empirical error is well calibrated in terms of ENCE and SSE but the uncertainty estimates are clearly not very useful.
In addition to being calibrated, it is generally desirable for uncertainty estimates to be as small as possible and to have some variation.
This characteristic is often referred to as \emph{sharpness} (or \emph{refinement})~\cite{tran2020, kuleshov2018, scalia2020, degroot1983}.
To evaluate the sharpness of a regression model we compute the root mean predicted variance (RMV) on unseen data.
A low RMV indicates the model on average predicts low uncertainty and thus low expected error.
Additionally, we compute the coefficient of variation (CV)~\cite{levi2020} of the predicted uncertainties on unseen data:
\begin{equation}
    \text{CV} = \frac{\sqrt{\tfrac{1}{N} \sum_{n=1}^N (\sigma_{*}(g_n) - \overline\sigma_*)^2}}{\overline\sigma_*} \, ,
\label{eq:coefficient_of_variation}
\end{equation}
where $\sigma_{*}(g_n)$ is the predicted standard deviation (uncertainty) of instance $n$,
$\overline\sigma_* = \frac{1}{N} \sum_{n=1}^N \sigma_*(g_n)$
is the mean predicted standard deviation and $N$ in this case iterates the test set.
A high CV indicates large dispersion (heteroscedasticity) and thus a high input dependence of the uncertainty estimates, whereas a CV of zero indicates constant (homoscedastic) and thus uninformative uncertainty estimates.

\subsection{Uncertainty recalibration}
\label{sec:uncertainty_calibration}

Often the training of machine learning models does not ensure calibration when the models are presented with unseen data.
Thus there is a need to recalibrate the predictive distribution to unseen data \emph{post hoc}, which can be achieved by applying a recalibration function, that maps the uncalibrated predictive distribution to a well calibrated distribution.
In our case, training each model with NLL loss can result in overfitting of the uncertainty to the training data
resulting in overconfident predictions on unseen data~\cite{nix}.
On the other hand, applying an ensemble approximation is expected to reduce the overall error, and should thus lead to lower uncertainty.
This is not reflected in the ensemble variance (equation~\ref{eq:ensemble_variance}) which is strictly higher than the average of the individual variances.
Furthermore, there is nothing in the training procedure which ensures that the ensemble variance (epistemic uncertainty) fits the error distribution.

Several approaches to \emph{post hoc} recalibration of regression models have been proposed in the literature~\cite{kuleshov2018, song2019, levi2020, musil2019}.
A straightforward, yet robust, method is to simply scale the predicted uncertainty estimates by a scaling factor $s_n^2$
optimised to minimise the NLL on a held out calibration dataset~\cite{levi2020, musil2019}, which has the advantage that it does not influence the mean prediction $\mu_*(g_n)$ and the calibrated predictive distribution remains a normal distribution:
\begin{equation}
p_{*s^2}(y_n|g_n) = \mathcal{N}\big( \mu_*(g_n), s_n^2 \sigma_*^2(g_n) \big) \, .
\end{equation}
In the simplest case, all uncertainty estimates are scaled by the same scaling factor, however, we achieved better results by applying a non-linear scaling function.
Specifically, to obtain the scaled uncertainty estimates we apply an isotonic regression model\footnote{We use the implementation of isotonic regression available from the scikit-learn Python package~\cite{scikit-learn}: \texttt{sklearn.isotonic.IsotonicRegression}.}  $f_\phi(\cdot)$
to fit the empirical squared errors $(y_n-\mu_\theta(g_n))^2$ on a held out calibration dataset:
\begin{equation}
s_n^2\sigma_*^2(g_n) = f_\phi(\sigma_*^2(g_n)) \quad \Leftrightarrow \quad s_n^2 = \frac{f_\phi(\sigma_*^2(g_n))}{\sigma_*^2(g_n)} \, .
\end{equation}
Thus, the recalibration function $f_\phi(\cdot)$ takes as input the uncalibrated uncertainty $\sigma_*^2(g_n)$ and outputs the scaled uncertainty $s_n^2\sigma_*^2(g_n)$.
The isotonic regression approach results in a monotonic increasing scaling function and thus has the desired property of being non-linear while maintaining the overall ordering of the uncertainty estimates.

\section{Experiments and results}
\label{sec:results}

\subsection{Datasets}

In our experiments we consider two publicly available datasets: QM9~\cite{qm9}, which is a widely used benchmark for machine learning predictions of molecular properties, and the more recent PC9~\cite{pc9}, that contains a more diverse set of molecules selected with the same general constraints as QM9.
The QM9 dataset consists of 133,885 small organic molecules in equilibrium state with up to 9 heavy atoms (C, O, N, F) besides hydrogen.
For each molecule, the dataset contains several quantum chemical properties calculated at the B3LYP/6-31G(2df,p) level of theory including total energy $U_0$, which incorporates the vibrational zero point energy (ZPE)~\cite{qm9}.
We additionally compute the total energy without the ZPE, $E = U_0 - \text{ZPE}$, to enable comparison with PC9, that does not include $U_0$.
The PC9 dataset~\cite{pc9} consists of 99,234 molecules extracted from the PubChem database~\cite{pubchem} by applying the constraints of QM9 outlined above and was found to represent a more diverse set of molecules than QM9.
PC9 includes properties calculated at the B3LYP/6-31G(d) level of theory including total energy $E$.
Structures that appear in both datasets were identified by comparing International Chemical Identifiers (InChI)~\cite{inchi} (see supplementary material~\ref{sec:inchi_comparison_details} for details).
We found that 21,777 molecules from QM9 are also in PC9 and 21,619 molecules from PC9 are also in QM9 (since QM9 contains duplicate InChi strings the numbers are not identical).
In line with previous work, we consider the atomisation energies (the energy remaining after subtracting the energies of the constituent atoms) in our experiments, rather than the actual total energies.
Thus in subsequent sections, $U_0$ and $E$ will be used to refer to the respective atomisation energies.

\subsection{Experimental setup}

To evaluate the proposed method, we performed computer experiments of predicting atomisation energies on the QM9 and PC9 datasets.
In each experiment, we trained an ensemble of $M=5$ message passing neural network models extended to predict uncertainty as described in section~\ref{sec:method}.
The models were trained individually using the same hyperparameters and data splits, but with random parameter initialisation and random shuffling of the training data to induce model diversity.
Following previous work~\cite{jorgensen}, the networks were constructed with $T=3$ interaction steps, a cutoff distance of 5.0~\AA~for generating the molecular graphs, and an embedding size of 256.
We used the PyTorch implementation of the AdamW optimizer~\cite{adamw} with an initial learning rate of 0.0001, an exponential decay learning rate scheduler, and a weight decay coefficient of 0.01.
Each model was trained for up to 3,000,000 gradient steps with a batch size of 100. 
The first 1,000,000 steps were used for warmup training using only MSE loss ($\lambda=1$) and then the loss was interpolated linearly from MSE to NLL on the next 1,000,000 steps (see equation \ref{eq:loss_interpolation}).
The validation set was used for early stopping with NLL criterion and was also used as calibration set for fitting the recalibration function $f_\phi$ as described in section~\ref{sec:uncertainty_calibration}.

\begin{table}[t]
\footnotesize
\centering
\begin{tabular}{lllrrrrrrr}
\toprule
\multicolumn{3}{c}{Dataset} & \multicolumn{2}{c}{Error (eV)} & \multicolumn{3}{c}{Calibration} & \multicolumn{2}{c}{Sharpness} \\
\cmidrule(r){1-3}\cmidrule(lr){4-5}\cmidrule(lr){6-8}\cmidrule(l){9-10}
 Train & Test & $y$ & MAE & RMSE & NLL & ENCE & SSE & RMV & CV \\ 
\midrule
QM9 & QM9 & $U_0$ & 0.0094 & 0.0313 & -3.1593 & 0.0484 & 0.0958 & 0.0275 & 1.8939 \\
QM9 & QM9 & $E$ & 0.0101 & 0.0342 & -3.0759 & 0.0720 & 0.1083 & 0.0270 & 1.7293 \\
PC9 & PC9 & $E$ & 0.0199 & 0.0844 & -2.5956 & 0.0650 & 0.1177 & 0.0612 & 2.2011 \\
QM9 & PC9 & $E$ & 0.4192 & 0.7410 &  0.8107 & 0.0220 & 0.4129 & 0.7441 & 0.6294 \\
PC9 & QM9 & $E$ & 0.1165 & 0.1737 & -0.5366 & 0.0312 & 0.0175 & 0.1781 & 0.5597 \\
\bottomrule
\end{tabular}
\caption{
Test results of ensemble models ($M=5$) trained to predict atomisation energy properties on the QM9 and PC9 datasets.
Mean absolute error (MAE) and root mean squared error (RMSE) are presented in electron volt (eV).
The uncertainty calibration in each experiment is summarised by the mean negative log likelihood (NLL), expected normalised calibration error (ENCE), and sum of squared errors (SSE).
The  uncertainty sharpness is summarised by the root mean variance (RMV) and coefficient of variation (CV) of the predicted uncertainties.
}
\label{tab:results}
\end{table}

\subsection{Prediction of $U_0$ on QM9 with random split}

\begin{figure}[t]
    \centering
    \begin{subfigure}[b]{0.49\textwidth}
        \centering
        \includegraphics[height=3.5cm]{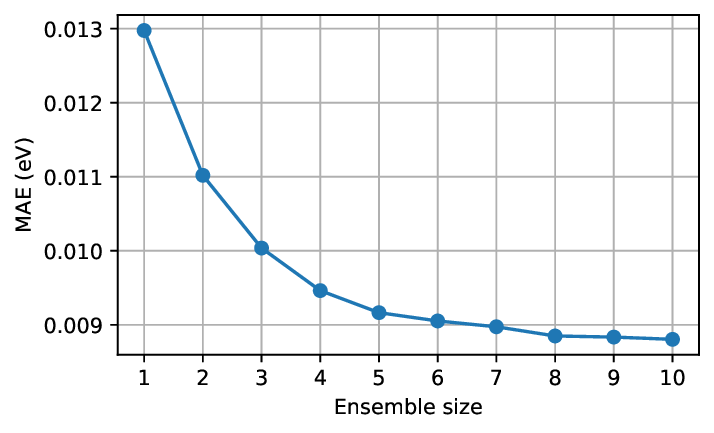}
        \caption{}
    \end{subfigure}
    \begin{subfigure}[b]{0.49\textwidth}
        \centering
        \includegraphics[height=3.5cm]{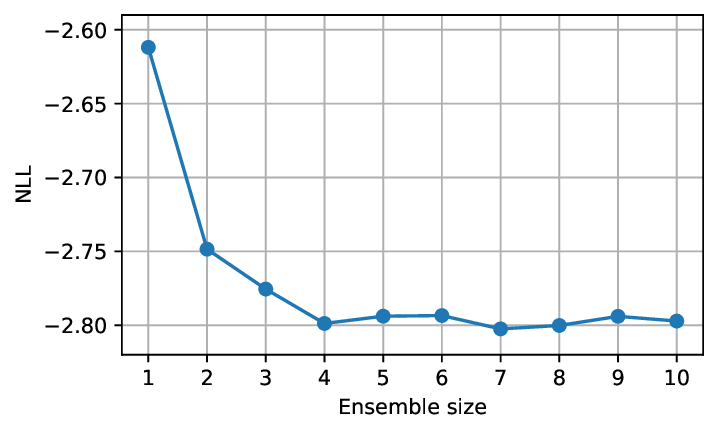}
        \caption{}
    \end{subfigure}
    \caption{Trade off between error and ensemble size evaluated on the QM9 validation set when predicting atomisation energy $U_0$:
    (a) mean absolute error (MAE) measured in eV and (b) mean negative log likelihood (NLL).
    The models are sorted by NLL in increasing order (best first).
    Reasonably low errors can be achieved with an ensemble size of $M=5$ models.}
    \label{fig:ensemble_size}
\end{figure}

In this first experiment, we trained an ensemble to predict the atomisation energy $U_0$ of the QM9 dataset.
Following previous work~\cite{messagepassing, jorgensen, schnet}, we randomly split the data into a training set of 110,000 molecules, a validation set of 10,000 molecules, and a test set consisting of the remaining 13,885 molecules.
Figure~\ref{fig:ensemble_size} shows the trade off between error and ensemble size of up to $M=10$ models on the validation set.
As expected, using a larger ensemble reduces the error, however, a reasonably low error was achieved with an ensemble of $M=5$ models and not much is gained beyond that, so we choose to use ensembles of this size throughout our experiments.
The test set results for an ensemble of $M=5$ models are presented in the first row of table~\ref{tab:results}.
The ensemble achieved a MAE~=~0.0094 eV
which is comparable to previous work using a similar model~\cite{jorgensen} (MAE~=~0.0105 eV), which indicates we did not lose any accuracy by extending the model to predict uncertainty.

\begin{figure}[t]
    \centering
    \begin{subfigure}[b]{0.32\textwidth}
        \centering
        \includegraphics[height=3.6cm]{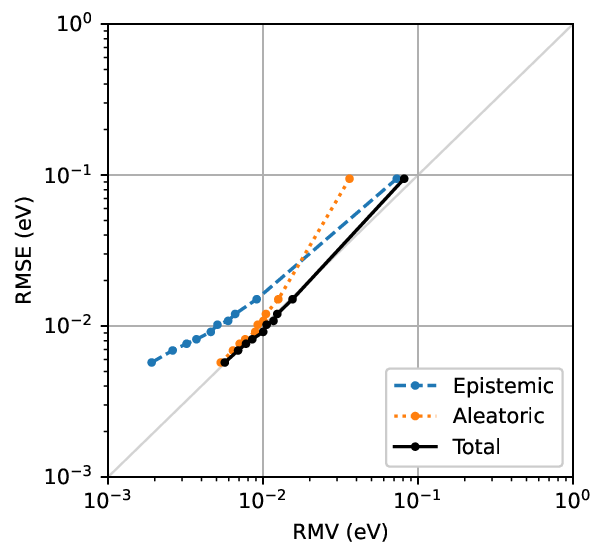}
        \caption{}
        \label{fig:uncertainty_evaluation_qm9U0_a}
    \end{subfigure}
    \begin{subfigure}[b]{0.32\textwidth}
        \centering
        \includegraphics[height=3.6cm]{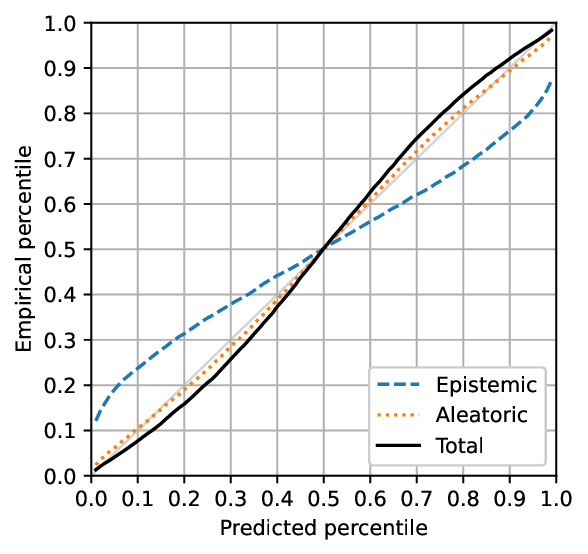}
        \caption{}
        \label{fig:uncertainty_evaluation_qm9U0_b}
    \end{subfigure}
    \begin{subfigure}[b]{0.32\textwidth}
        \centering
        \includegraphics[height=3.6cm]{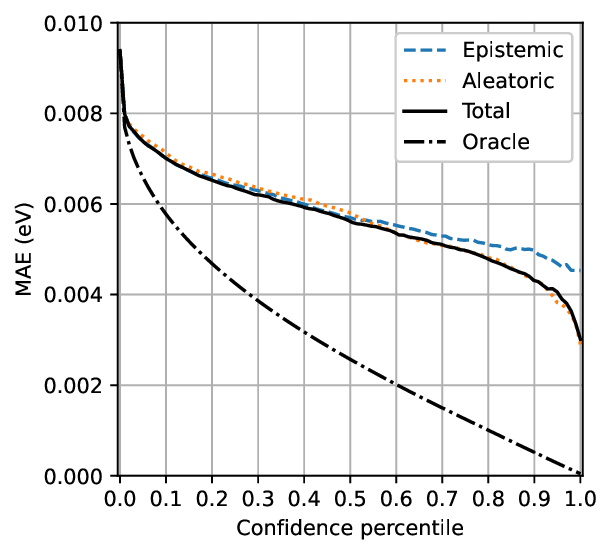}
        \caption{}
        \label{fig:uncertainty_evaluation_qm9U0_c}
    \end{subfigure}
    \caption{Evaluation of uncertainty on the QM9 test set when predicting atomisation energy $U_0$.
    The error-calibration plot (a) shows empirical root mean squared error (RMSE) as a function of predicted uncertainty measured by root mean variance (RMV) computed in bins.
    The quantile-calibration plot (b) compares predicted percentiles and empirical percentiles averaged over the test data. 
    The confidence curve (c) shows the variation in mean absolute error (MAE) as a function of the uncertainty threshold.
    }
    \label{fig:uncertainty_evaluation_qm9U0}
\end{figure}

After training the ensemble model, the ensemble predictive distribution was recalibrated by fitting an isotonic regression recalibration function (see section~\ref{sec:uncertainty_calibration}) on the validation set and applying it on the test set resulting in an average scaling factor of 0.2965 (SD~=~0.5346) on the test set (where SD denotes the standard deviation).
Even though each individual model in the ensemble is expected to have increased error when presented with unseen data, the ensemble approximation significantly improved the overall error in this case resulting in a recalibration function that effectively shrinks the uncertainty of the predictive distribution.
Uncertainty calibration plots are presented in figure~\ref{fig:uncertainty_evaluation_qm9U0} and uncertainty calibration and sharpness metrics are included in the first row of table \ref{tab:results}.
The error-calibration plot (figure~\ref{fig:uncertainty_evaluation_qm9U0_a}) shows that in general the model assigns higher uncertainty to instances with higher error as desired.
Hence the uncertainty estimates are highly input dependent and have high dispersion as also indicated by a high CV.
Overall the model is well calibrated in terms of error-calibration since the predicted uncertainties correspond closely to the expected empirical errors on average resulting in a low ENCE.
The rightmost point in the plot, representing the bin with the highest uncertainty estimates, includes instances with relatively large errors, placing this point far from the rest.
However, the model correctly assigns high uncertainty to these instances, thereby identifying them as problematic.
The error-calibration plot also reveals that for low uncertainty predictions the epistemic uncertainty is relatively low, indicating a high level of agreement among the individual models of the ensemble, and consequently the aleatoric uncertainty is responsible for the majority of the total uncertainty in these cases.
On the other hand, the high uncertainty predictions have relatively high epistemic uncertainty, corresponding to a high level of disagreement among the individual models, indicating these molecules are out of distribution and therefore the predictions are also more likely to have high error.
The quantile-calibration plot (figure~\ref{fig:uncertainty_evaluation_qm9U0_b}) shows that the percentiles of the predicted distributions corresponds well to the empirical distribution on average resulting in a low SSE, and the symmetry at the 0.5 percentile indicates that the error distribution is not skewed and the model is not biased.
In the confidence curve (figure~\ref{fig:uncertainty_evaluation_qm9U0_c}), the downwards slope indicates that the uncertainty estimates provide a meaningful ranking of the predictions with respect to the error.
Interestingly, leaving out the 10\% most uncertain predictions results in a significant decrease in error, indicating a potentially large benefit from including these molecules in the training data to improve the error on similar examples in the future following an active learning methodology.
Considering only the most confident predictions results in a lower average error as desired.


\begin{figure}[t]
    \centering
    \begin{subfigure}[b]{0.49\textwidth}
        \centering
        \includegraphics[height=3.5cm]{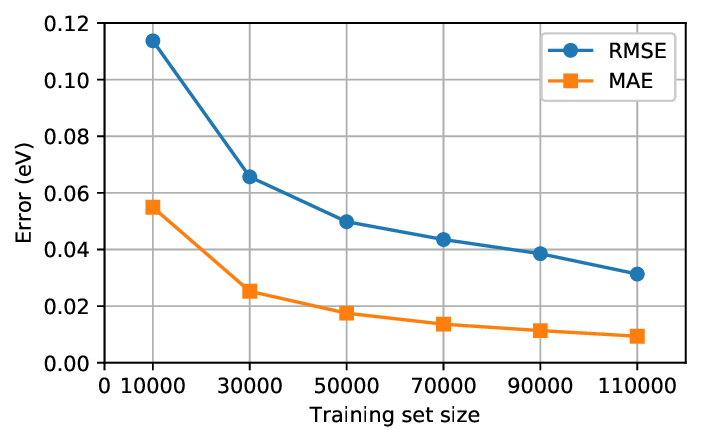}
        \caption{}
    \end{subfigure}
    \begin{subfigure}[b]{0.49\textwidth}
        \centering
        \includegraphics[height=3.5cm]{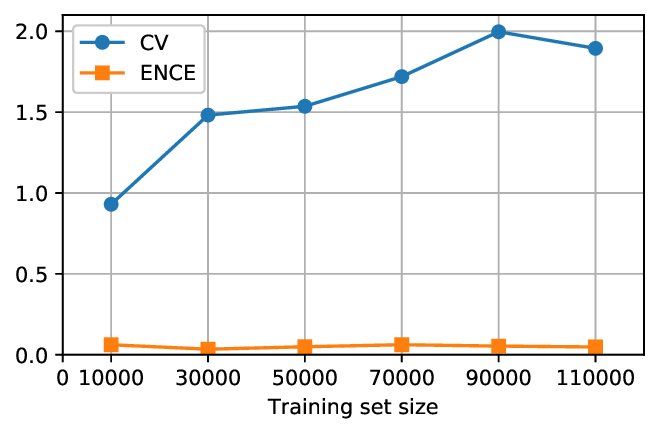}
        \caption{}
    \end{subfigure}
    \caption{Learning curves showing test set metrics as a function of training set size on the QM9 dataset when predicting $U_0$.
    (a) The mean absolute error (MAE) and root mean squared error (RMSE) improve with more data as expected.
    (b) The calibration in terms of expected normalised calibration error (ENCE) does not vary significantly, while the dispersion of uncertainty estimates measured by the coefficient of variation (CV) increases with the amount of training data.
    }
    \label{fig:learning_curve}
\end{figure}

Learning curves for this experiment are presented in figure~\ref{fig:learning_curve} showing test set metrics as a function of the amount of training data when predicting $U_0$ on QM9.
As expected, the errors decrease with more training data.
Interestingly, good calibration in terms of the ENCE was obtained with relatively small training datasets and the ENCE does not vary significantly when adding more data, while the sharpness of uncertainty estimates measured by the CV clearly increases with the amount of training data, making the uncertainty estimates more input dependent and thus more informative.

\subsection{Prediction of $E$ on QM9 with random split}

Complementary to the first experiment, we trained an ensemble to predict the atomisation energy $E$ of the QM9 dataset using the same data split.
This allows for more direct comparison with results from subsequent experiments using the PC9 dataset.
The test set results are presented in the second row of table~\ref{tab:results}.
The ensemble model achieved a MAE = 0.0101 eV, which is a little higher than when predicting $U_0$, indicating that predicting $E$ is slightly harder.
A similar finding was reported in~\cite{pc9} using a SchNet~\cite{schnet} model.

The uncertainty estimates were likewise recalibrated by fitting a recalibration function on the validation set and applying it on the test set resulting in an average scaling factor of 0.3116 (SD = 0.3966) on the test set, effectively shrinking the predictive distribution similarly to the first experiment.
Uncertainty calibration plots for this experiment are included in the supplementary material in figure~\ref{fig:uncertainty_evaluation_qm9E}.
As in the first experiment, we found that the model succeeds at assigning uncertainty estimates that correlates with the expected error and the model is well calibrated in terms of ENCE and SSE.

\subsection{Prediction of $E$ on PC9 with random split}

Next, we trained an ensemble to predict the atomisation energy $E$ of the more diverse PC9 dataset.
The data was split randomly into a training set of 80,000 molecules, a validation set of 10,000 molecules, and a test set consisting of the remaining 9,234 molecules.
The test set results are presented in the third row of table~\ref{tab:results}.
The ensemble model achieved a MAE = 0.0199 eV which is approximately twice as high as when predicting $E$ on QM9.
We attribute this increase in error to PC9 representing a more diverse set of molecules, making the task more difficult, and additionally to the smaller size of the training dataset.
A similar increase in error between QM9 and PC9 was reported in~\cite{pc9} using a SchNet~\cite{schnet} model.

The uncertainty estimates were recalibrated by fitting a recalibration function on the validation set and applying it on the test set resulting in an average scaling factor of 0.2938 (SD = 0.6449) on the test set, effectively shrinking the uncertainty of the predictive distribution similarly to the two previous experiments.
Uncertainty calibration plots for this experiment are included in the supplementary material in figure~\ref{fig:uncertainty_evaluation_pc9E}.
The model succeeds in assigning uncertainty estimates that correlates with the expected error and the model is well calibrated in terms of ENCE and SSE.
As in the two previous experiments, some instances among the predictions with the highest uncertainty have relatively large errors and account for a large part of the overall error as shown by the error-calibration plot.
The confidence curve shows that leaving out the 20\% most uncertain predictions almost halves the MAE, indicating a good ranking ability in this experiment. 

\subsection{Generalisation from QM9 to PC9}

In this experiment we examine the effect of testing an ensemble of models trained on QM9 on the more diverse set of molecules found in PC9 and especially how it affects the uncertainty estimates as we anticipate larger errors.
When constructing the data splits, we utilize the fact that the datasets are overlapping, using the 112,108 molecules that are unique to QM9 as the training set and the 21,777 structures from QM9 that are also present in PC9 as the validation set. 
Then we compute a linear correction on the 21,619 structures from PC9 that are present in QM9 to account for the different level of theory used to calculate the energy properties.
Following~\cite{pc9}, the linear correction was performed by fitting a Huber regression model (coefficient~=~1.0038, intercept~=~1.1428) on the predicted and observed energies.
Finally, we use the remaining 77,615 molecules exclusive to PC9 as the the test set and apply the linear correction to the predictions.
The test set results are presented in the fourth row of table~\ref{tab:results}.
The ensemble model achieved a MAE = 0.4192 eV, which is comparable to the findings reported in~\cite{pc9} using a SchNet~\cite{schnet} model. 
The relatively high error is caused primarily by out of distribution instances, and indicates that the model has problems generalising under domain shift, and secondly by the different level of theory used to calculate the energies in the two datasets, which was shown to produce large errors (see figure 3 in~\cite{pc9} for details).

As in the previous experiments, the predictive distribution was recalibrated by fitting a recalibration function on the validation set and applying it on the test set resulting in an average scaling factor of 135.0313 (SD = 31.5991) on the test set.
The large average scaling factor reflects the large increase in error caused by the more diverse dataset and different level of theory used to calculate the energies as mentioned above.
Uncertainty calibration figures for this experiment are presented in the supplementary material figure~\ref{fig:uncertainty_evaluation_qm9pc9E}.
Interestingly, the uncertainty estimates produced by the model are still well calibrated in terms of error-calibration as indicated by the low ENCE and thus the model correctly assigns high uncertainty to instances with large errors as desired.
The error-calibration plot additionally shows a larger contribution of the epistemic uncertainty to the total uncertainty in more cases compared to the other experiments, confirming that many of the examples are regarded as out of distribution by the model as hypothesised.
The quantile-calibration plot and the relatively high SSE shows that the predicted percentiles do not fit the empirical percentiles averaged over the dataset in this experiment.
This is primarily because the errors are not normally distributed in this particular case as was also reported in~\cite{pc9}.
As illustrated by the confidence curve, the uncertainty estimates provides a good ranking with respect to error among the high uncertainty estimates.
However, among the low uncertainty estimates there is little variation in the predicted uncertainties and the ranking is therefore uninformative resulting in a flat confidence curve.
The lack of variation in the uncertainty estimates also results in low sharpness in terms of CV.

\subsection{Generalisation from PC9 to QM9}

Now going in the opposite direction, in this last experiment we examine the effect of applying an ensemble of models trained on PC9 to the less diverse set of molecules in QM9.
Analogous to the previous experiment, we use the 77,615 molecules that are unique to PC9 as the training set and the 21,619 structures from PC9 that are also present in QM9 as the validation set.
Similarly to the previous experiment, we compute a linear correction on the 21,777 structures from QM9 that are also present in PC9 by fitting a Huber regression model (coefficient = 0.9994, intercept = \textminus0.6830) on the predicted and observed energies.
Finally, we use the remaining 112,108 molecules exclusive to QM9 as the test set.
The test set results are presented in the fifth and final row of table~\ref{tab:results}.
The ensemble achieved a MAE~=~0.1165 eV, which is comparable to the findings reported in~\cite{pc9} using a SchNet~\cite{schnet} model.
While high compared to the experiment of predicting $E$ on QM9 above, the error is significantly lower than the previous experiment of training on QM9 and testing on PC9 as might be expected when going from a more diverse dataset to an overlapping and less diverse dataset.
Some of the error may be attributed to the different level of theory used to calculate the energies in QM9 and PC9, respectively.

The uncertainty estimates were recalibrated by fitting a recalibration function on the validation set and applying it on the test set resulting in an average scaling factor of 6.2404 (SD = 1.4109) on the test set, which like the error is also significantly lower than the previous experiment.
Uncertainty calibration figures for this experiment are included in the supplementary material~\ref{fig:uncertainty_evaluation_pc9qm9E}.
Similarly to the previous experiment, the uncertainty is well calibrated in terms of error-calibration shown by a low ENCE.
However, in this experiment less of the total uncertainty is contributed to the epistemic uncertainty, indicating most cases are not regarded as out of distribution by the model as hypothesised.
In this case the uncertainty is also well calibrated in terms of quantile-calibration indicated by a low SSE further indicating there are not as many out of distribution examples.
While the model is well calibrated, there is less variation in the uncertainty estimates in this case resulting in a low CV.
The lack of sharpness gives the model a poor ranking ability compared to the other experiments as shown by the less steep slope of the confidence curve.

\section{Discussion}
\label{sec:discussion}

Through five computer experiments we have shown that the proposed ensemble approximation and recalibration method achieves good accuracy and uncertainty calibration on two publicly available benchmark datasets for molecular property prediction.
In the first three experiments, random data splits were used to train ensemble models to predict atomisation energies on the QM9 and PC9 datasets, respectively.
The result of predicting energy $U_0$ on QM9 is comparable with previous work by the authors using the same base model~\cite{jorgensen}, meaning we did not loose accuracy by extending the model to include predictive uncertainty.
We saw a small increase in the error when predicting $E$ on QM9 which is consistent with results reported in~\cite{pc9}.
The error when predicting $E$ on the more diverse PC9 dataset was almost twice as high compared to QM9, which is also consistent with results reported in~\cite{pc9}, indicating that the additional chemical diversity observed in this dataset makes the prediction task harder.
In all three random split experiments, the proposed method produced well calibrated uncertainty estimates characterised by highly correlated average uncertainties and errors as well as highly correlated predicted and empirical quantiles, as shown in the calibration plots in figure~\ref{fig:uncertainty_evaluation_qm9U0} and additionally in the corresponding figures in supplementary material~\ref{sec:additional_uncertainty_results}, and further summarized by low ENCE and SSE values presented in table~\ref{tab:results}.
The error-calibration plots further show that for the test examples with high error the epistemic uncertainty is high relative to the aleatoric uncertainty, indicating high variance among the predictions of the individual models in the ensemble.
This means that the ensemble model is good at identifying instances that are out of distribution and therefore have high expected error, and exemplifies why it is useful to be able to distinguish between epistemic and aleatoric uncertainty in the predictions.
In addition to being well calibrated, the uncertainty estimates were also sharp, as shown by combined low RMV and high CV values, indicating the predicted uncertainty estimates are highly input dependent and thereby informative.

In the fourth experiment, we aimed to generalise from QM9 to the more diverse PC9 dataset by training on QM9 and testing on molecules exclusive to PC9.
The analysis of the PC9 structures presented in~\cite{pc9} showed that some molecules included in PC9 are chemically different from molecules in QM9, making this experiment a difficult out of distribution prediction task.
Additionally, the properties of the datasets where computed at different levels of theory (B3LYP/6-31G(2df,p) in QM9 and B3LYP/6-31G(d) in PC9), which we accounted for with a linear correction, following~\cite{pc9}.
The error we observed in this experiment was quite high, but comparable to what is reported in~\cite{pc9}.
Importantly, the uncertainty estimates of our model were still well error-calibrated, meaning the model correctly identified the high error instances by assigning them high uncertainty, which means the out of distribution cases can be detected and handled.
The error-calibration plot (figure~\ref{fig:uncertainty_evaluation_qm9pc9E}a) shows that epistemic uncertainty was responsible for the majority of the total uncertainty in the high error cases in this experiment, correctly identifying these cases as problematic and out of distribution.
The model does not have good quantile-calibration since the errors in this experiment are not normally distributed as also shown in~\cite{pc9}.
In the fifth and final experiment, we went in the opposite direction and trained on PC9 to predict the molecules exclusive to QM9. 
This should be an easier task, since QM9 is similar to but less diverse than PC9. 
As expected, the error we observed is significantly lower than in the previous experiment and comparable to what was reported in~\cite{pc9}.
The model produced well calibrated uncertainty estimates in terms of both error- and quantile-calibration but achieved poor sharpness, which means the uncertainty estimates were less informative in this case.
Figure~\ref{fig:learning_curve} indicates that perhaps better sharpness can be achieved with more training data.
Interestingly, the two generalisation experiments resulted in the best overall error-calibration of all the experiments in terms of ENCE despite having the largest errors (see table~\ref{tab:results}).
They also achieved the poorest sharpness measured by CV.
Furthermore, in the learning curve experiment presented in figure~\ref{fig:learning_curve} we observed that good calibration was achieved even for small training set sizes where the error is relatively high and that sharpness seems to increase with the amount of training data.
This illustrates how calibration is orthogonal to accuracy~\cite{deepensemble} and further shows the importance of measuring sharpness in addition to calibration to ensure uncertainty estimates are not only well calibrated but also informative.

The effectiveness of the ensemble approximation in the proposed method, and thus the quality of the epistemic uncertainty estimates, depends on training a diverse set of models to ensure variance of predictions beyond the training data distribution.
In this work we rely on random initialisation of network parameters and random shuffling of the training data to induce model diversity, but other more deliberate methods exist.
Bootstrapping, i.e.~re-sampling the training set with replacement, is a popular technique for inducing diversity in ensemble models, but some evidence suggests that this method is less appropriate for deep models as they typically perform better with more training data~\cite{deepensemble}.
We tried to apply bootstrapping in our experiments, but did not observe any improvements in terms of error or calibration, so we left it out for simplicity.
Another more recent approach to induce diversity is to use randomized prior functions~\cite{randomizedpriorfunctions}, which we consider an interesting direction for future work. 

A major advantage of the proposed method is its ability to quantify and distinguish between epistemic and aleatoric uncertainty in the predictions.
Both types of uncertainty are necessary to asses the total uncertainty and thus for obtaining well calibrated uncertainty estimates in and out of the training data distribution. 
Modelling aleatoric uncertainty explicitly is important for capturing heteroscedastic noise in the data and thereby making input dependent predictions of the noise wheres capturing epistemic uncertainty is especially important in tasks where it is expected to encounter out of distribution instances.
Chemical space is so vast that it is not feasible to gather enough training data to cover the entire domain~\cite{gomez-bombarelli, polishchuk2013}.
Thus, identifying cases beyond the training data distribution where the model is not expected to be accurate is critical.
For example, distinguishing between epistemic and aleatoric uncertainty can be utilised in a screening system for atomic structures. 
If the epistemic uncertainty of a prediction is low, the aleatoric uncertainty indicates the expected error.
If, on the other hand, the epistemic uncertainty is high, there is a high level of disagreement in the ensemble and therefore low confidence in the prediction, and the system can automatically fall back to a more accurate and computationally expensive method such as DFT~\cite{bhowmik2019}. 
In an active learning setting, the epistemic uncertainty is important for detecting out of distribution candidates that can be included in the training data to make the model generalise better on a wider domain.
The specific confidence thresholds for decision making can be tuned depending on the data, application and computational resources available.

\section{Conclusion}
\label{sec:conclusion}

In this work we have explored a complete framework for obtaining well calibrated uncertainty estimates for accurate molecular property prediction by using a deep ensemble of message passing neural networks and \emph{post hoc} recalibrating the uncertainty estimates to unseen data.
Our experiments on two publicly available benchmark datasets have showed that the method is able to produce well calibrated uncertainty estimates in and out of the training data distribution such that on average the model assigns high uncertainty to high error examples.
A major advantage of the proposed approach is that the uncertainty estimates can be decomposed into epistemic and aleatoric uncertainty, which provides important information for decision making, crucial in for example high throughput screening and active learning applications.
Additionally, the proposed method does not depend on the particular architecture of the neural network model, and can thus easily be adapted to use other domain-specific models and new improved models as research in model development advances.

\section{Acknowledgements}
The authors acknowledge support from the Novo Nordisk Foundation (SURE, NNF19OC0057822) and the European Union’s Horizon 2020 research and innovation program under grant agreement No 957189 (BIG-MAP) and No 957213 (BATTERY2030PLUS).

\bibliographystyle{unsrt}
\bibliography{references}

\clearpage
\input{appendix}

\end{document}

%% file: appendix.tex
\appendix

\setcounter{page}{1}

\section*{Supplementary material:}

{\bf Calibrated Uncertainty for Molecular Property Prediction\\using Ensembles of Message Passing Neural Networks}

\section{InChi comparison details}
\label{sec:inchi_comparison_details}

The overlapping set of structures that appear in both the QM9~\cite{qm9} and PC9~\cite{pc9} datasets were identified by comparing International Chemical Identifiers (InChI) strings~\cite{inchi}.
The InChI strings for both datasets were computed using the Open Babel command line tool (obabel v. 3.1.0):
\begin{verbatim}
$ obabel [input_file.xyz] -o inchi -xr -O [output_file.inchi]
\end{verbatim}
\noindent or similarly for multiple files:
\begin{verbatim}
$ for f in *.xyz;
> do obabel $f -o inchi -xr -O ../inchi/${f:0:-3}inchi;
> done
\end{verbatim}
Then the InChi strings were truncated as to not differentiate between stereoisomers (structures with the same chemical formula and connectivity).
Specifically, the {\verb /b }, {\verb /t }, {\verb /m }, and {\verb /s } layers of the InChi strings were removed.
When comparing the truncated InChi strings of the two datasets, we found that that 21,777 molecules from QM9 are also in PC9 and 21,619 molecules from PC9 are also in QM9.
The numbers are not identical since QM9 and PC9 contains a different amount of duplicate truncated InChi strings, so a structure from one dataset can appear multiple times in the other dataset.

In~\cite{pc9} it was reported that 18,357 structures from PC9 also belong to QM9, based on comparing InChi strings computed with the Open Babel software.
We were not able to reproduce this number using any combination of InChi layers, so we instead used the method and result described above in this section.

\section{Additional results}
\label{sec:additional_uncertainty_results}

\renewcommand{\thefigure}{B\arabic{figure}}
\setcounter{figure}{0}

\begin{figure}[H]
    \centering
    \begin{subfigure}[b]{0.32\textwidth}
        \centering
        \includegraphics[height=3.6cm]{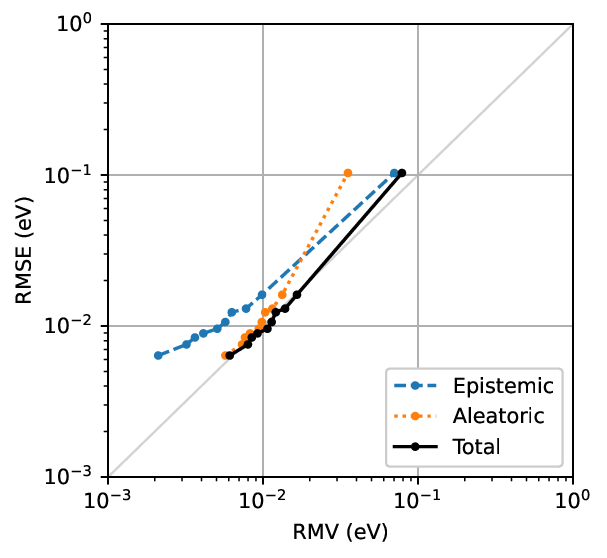}
        \caption{}
    \end{subfigure}
    \begin{subfigure}[b]{0.32\textwidth}
        \centering
        \includegraphics[height=3.6cm]{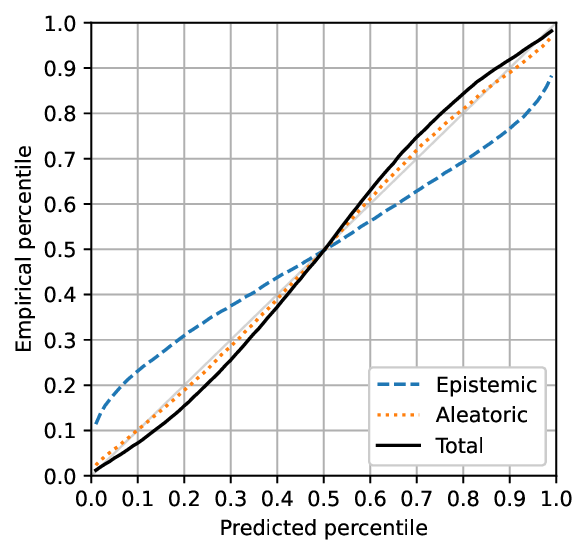}
        \caption{}
    \end{subfigure}
    \begin{subfigure}[b]{0.32\textwidth}
        \centering
        \includegraphics[height=3.6cm]{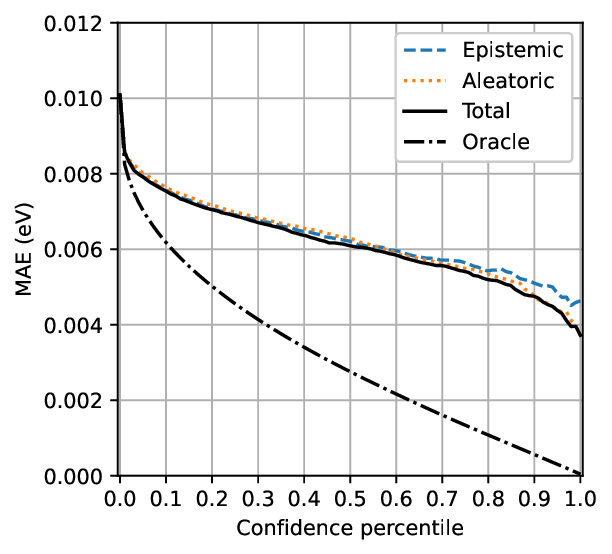}
        \caption{}
    \end{subfigure}
    \caption{Evaluation of uncertainty on the QM9 test set when predicting $E$:
    (a) error-calibration plot, (b) quantile-calibration plot, and (c) confidence curve.}
    \label{fig:uncertainty_evaluation_qm9E}
\end{figure}

\begin{figure}[H]
    \centering
    \begin{subfigure}[b]{0.32\textwidth}
        \centering
        \includegraphics[height=3.6cm]{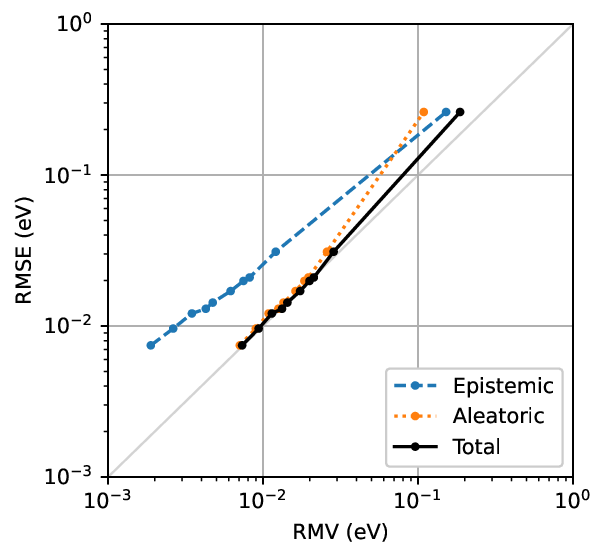}
        \caption{}
    \end{subfigure}
    \begin{subfigure}[b]{0.32\textwidth}
        \centering
        \includegraphics[height=3.6cm]{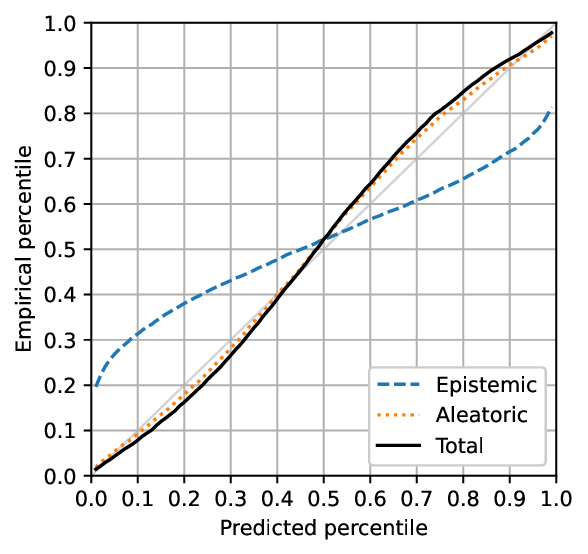}
        \caption{}
    \end{subfigure}
    \begin{subfigure}[b]{0.32\textwidth}
        \centering
        \includegraphics[height=3.6cm]{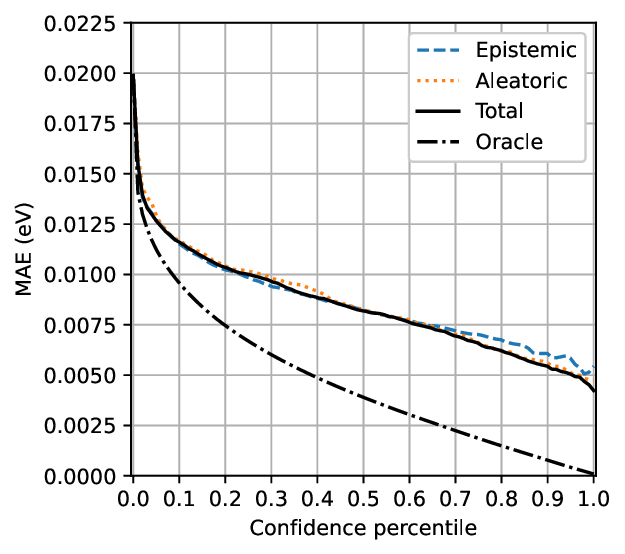}
        \caption{}
    \end{subfigure}
    \caption{Evaluation of uncertainty on the PC9 test set when predicting $E$:
    (a) error-calibration plot, (b) quantile-calibration plot, and (c) confidence curve.}
    \label{fig:uncertainty_evaluation_pc9E}
\end{figure}

\begin{figure}[H]
    \centering
    \begin{subfigure}[b]{0.32\textwidth}
        \centering
        \includegraphics[height=3.6cm]{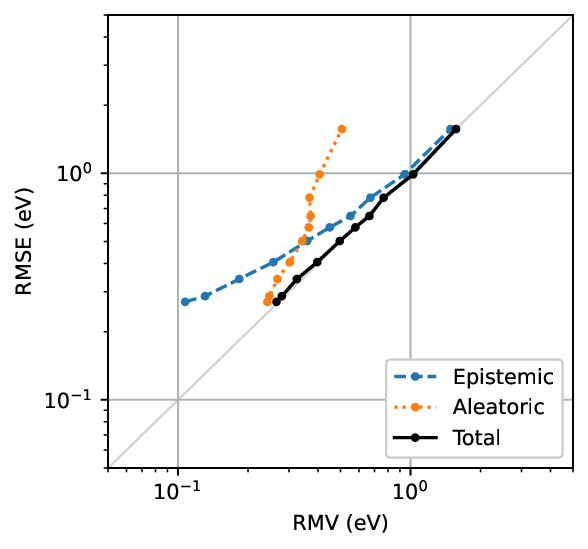}
        \caption{}
    \end{subfigure}
    \begin{subfigure}[b]{0.32\textwidth}
        \centering
        \includegraphics[height=3.6cm]{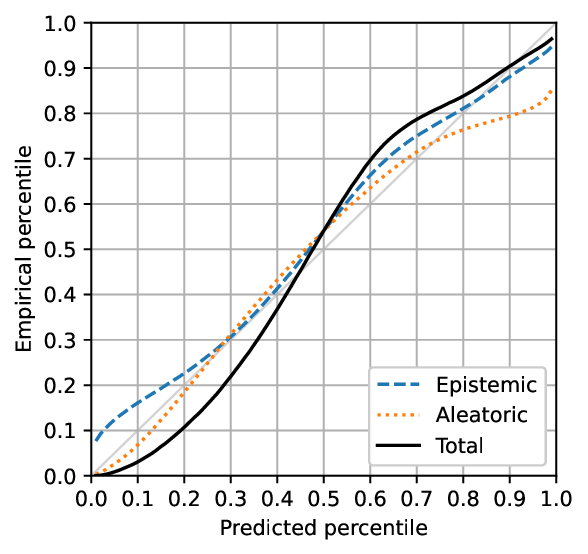}
        \caption{}
    \end{subfigure}
    \begin{subfigure}[b]{0.32\textwidth}
        \centering
        \includegraphics[height=3.6cm]{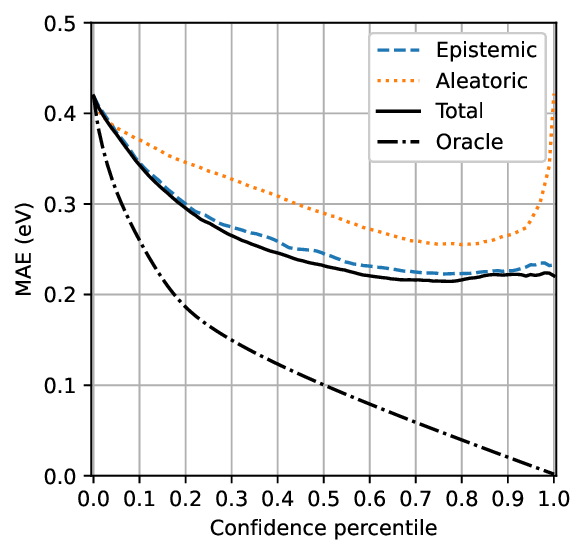}
        \caption{}
    \end{subfigure}
    \caption{Evaluation of uncertainty when training on QM9 and testing on PC9:
    (a) error-calibration plot, (b) quantile-calibration plot, and (c) confidence curve.}
    \label{fig:uncertainty_evaluation_qm9pc9E}
\end{figure}

\begin{figure}[H]
    \centering
    \begin{subfigure}[b]{0.32\textwidth}
        \centering
        \includegraphics[height=3.6cm]{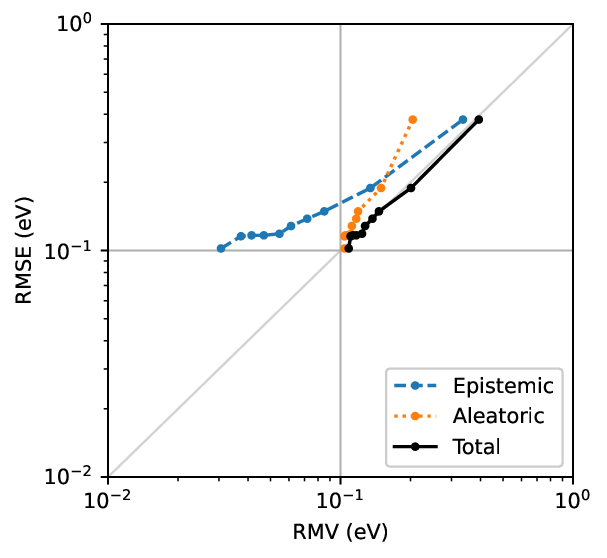}
        \caption{}
    \end{subfigure}
    \begin{subfigure}[b]{0.32\textwidth}
        \centering
        \includegraphics[height=3.6cm]{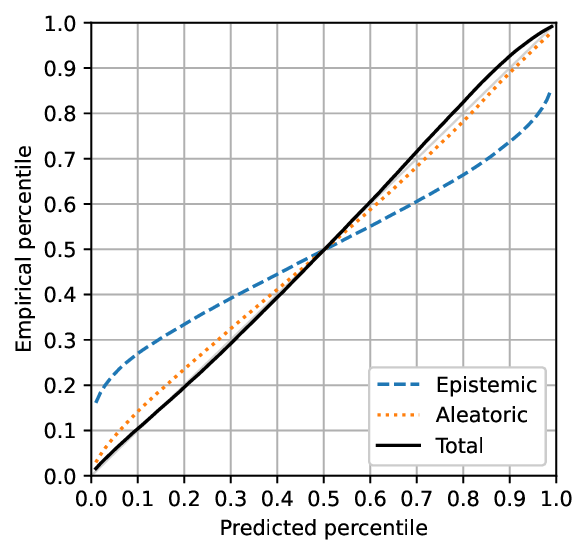}
        \caption{}
    \end{subfigure}
    \begin{subfigure}[b]{0.32\textwidth}
        \centering
        \includegraphics[height=3.6cm]{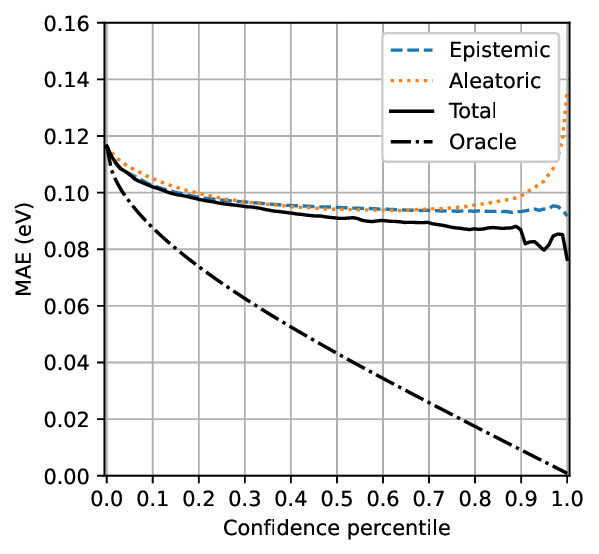}
        \caption{}
    \end{subfigure}
    \caption{Evaluation of uncertainty when training on PC9 and testing on QM9:
    (a) error-calibration plot, (b) quantile-calibration plot, and (c) confidence curve.}
    \label{fig:uncertainty_evaluation_pc9qm9E}
\end{figure}